\documentclass{article}

\usepackage[final]{neurips_2024}

\usepackage[utf8]{inputenc} 
\usepackage[T1]{fontenc}    
\usepackage{url}            
\usepackage{booktabs}       
\usepackage{amsfonts}       
\usepackage{nicefrac}       
\usepackage{microtype}      
\usepackage{xcolor}         
\usepackage{multirow}
\usepackage{makecell}
\usepackage{graphicx}
\usepackage{pifont}
\usepackage{colortbl}
\usepackage{natbib}
\usepackage[pagebackref=true,breaklinks=true,colorlinks,bookmarks=false]{hyperref}

\newlength\savewidth

\newcommand{\cmark}{\ding{51}}
\newcommand{\xmark}{\ding{55}}

\makeatletter

\newcommand{\Rmnum}[1]{\expandafter@slowromancap\romannumeral #1@}
\makeatother

\def\eg{\textit{e.g.}}
\def\ie{\textit{i.e.}}

\title{SlowFocus: Enhancing Fine-grained Temporal Understanding in Video LLM}

\author{
    \hspace{-12mm}
    Ming Nie$^1$
    \quad
    Dan Ding$^1$
    \quad
    Chunwei Wang$^2$
    \quad
    Yuanfan Guo$^2$
    \quad
    Jianhua Han$^3$
    \quad
    Hang Xu$^3$
    \quad
    Li Zhang$^1$\thanks{Li Zhang (lizhangfd@fudan.edu.cn) is the corresponding author.}\quad\quad\quad\quad
    \vspace{.3em} 
  \\
  \hspace{-27mm}
  $^1$School of Data Science, Fudan University 
  \quad
  $^2$Noah's Ark Lab, Huawei
  \\
  \hspace{-27mm}
  $^3$Yinwang Intelligent Technology Co., Ltd.
    \vspace{.5em}
  \\
  \hspace{-27mm}
  \url{https://github.com/fudan-zvg/SlowFocus}
}

\begin{document}

\maketitle
\begin{abstract}
Large language models (LLMs) have demonstrated exceptional capabilities in text understanding, which has paved the way for their expansion into video LLMs (Vid-LLMs) to analyze video data. 
However, current Vid-LLMs struggle to simultaneously retain high-quality frame-level semantic information (\ie, a sufficient number of tokens per frame) and comprehensive video-level temporal information (\ie, an adequate number of sampled frames per video).
This limitation hinders the advancement of Vid-LLMs towards fine-grained video understanding.
To address this issue, we introduce the SlowFocus mechanism, which significantly enhances the equivalent sampling frequency without compromising the quality of frame-level visual tokens.
SlowFocus begins by identifying the query-related temporal segment based on the posed question, then performs dense sampling on this segment to extract local high-frequency features.
A multi-frequency mixing attention module is further leveraged to aggregate these local high-frequency details with global low-frequency contexts for enhanced temporal comprehension.
Additionally, to tailor Vid-LLMs to this innovative mechanism, we introduce a set of training strategies aimed at bolstering both temporal grounding and detailed temporal reasoning capabilities.
%
Furthermore, we establish FineAction-CGR, a benchmark specifically devised to assess the ability of Vid-LLMs to process fine-grained temporal understanding tasks.
Comprehensive experiments demonstrate the superiority of our mechanism across both existing public video understanding benchmarks and our proposed FineAction-CGR.

\end{abstract}
\section{Introduction}
Large language models (LLMs) have garnered significant attention due to their exceptional text understanding capabilities.
Building on the strengths of LLMs, video large language models (Vid-LLMs)~\cite{li2023videochat, lin2023videollava, zhang2023videollama} adapt them to the video modality, extending their reasoning and interactive skills to video data.
By training on video-level tasks like captioning and question answering, they establish coarse-grained video-language correspondence and acquire the capabilities of video understanding.

\begin{figure*}[t]
    \begin{center}
        \includegraphics[width=1.0\linewidth]{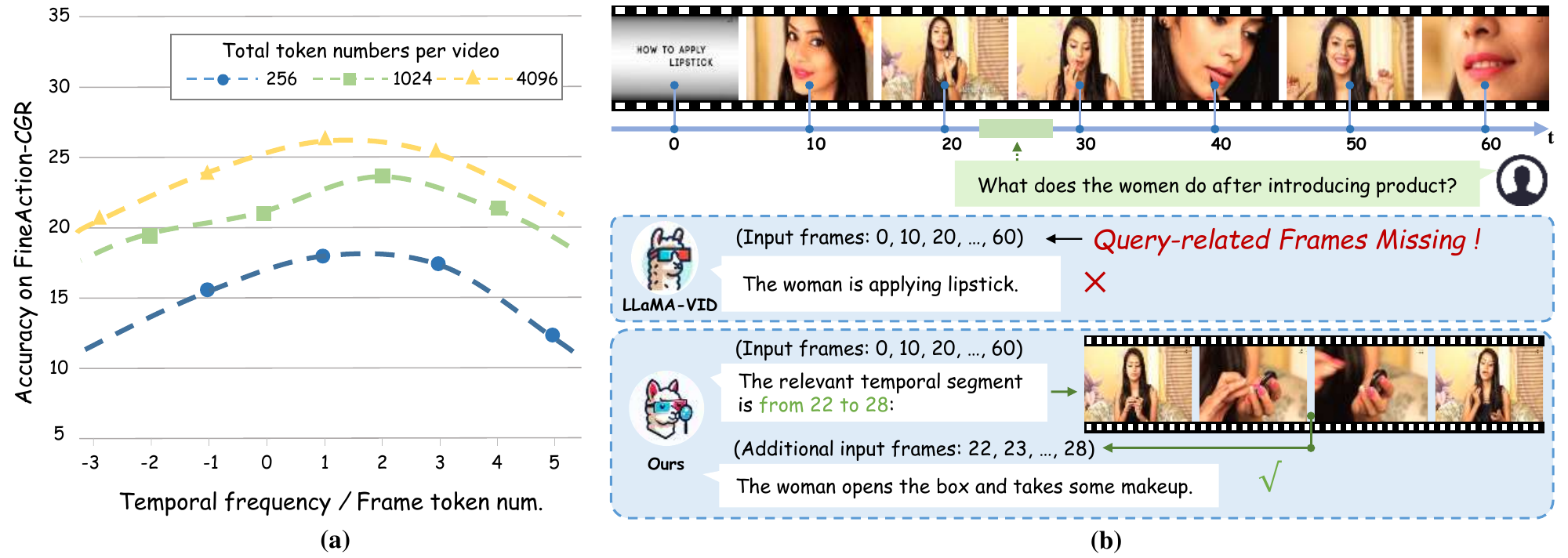}
    \end{center}
    \vspace{-3mm}
    \caption{(a) Trade-off between video sampling frequency and frame token number. The horizontal axis represents the ratio (log-transformed) of these two factors. Each curve corresponds to a fixed total number of tokens (\eg, 256 for a 1-minute video). (b) Deficiency of existing Vid-LLMs, such as LLaMA-VID, when facing fine-grained video understanding, and the efficacy of our approach.}
    \label{fig-intro}
    \vspace{-2mm}
\end{figure*}

While Vid-LLMs have demonstrated the promising performance in video understanding, they still face a significant challenge.
To embed video features into LLMs under computing resource constraints, Vid-LLMs typically need to sparsely sample the original video (\eg, retaining one frame every second) into a collection of low-frequency frames.
Subsequently, the token number of each frame are also compressed through a visual adapter like average pooling~\cite{li2023llama} or Q-former~\cite{li2023blip}.
This leads to a dilemma: Vid-LLMs have to choose between compromised frame-level features and video-level features, each resulting in a reduction of video information.
As illustrated in Figure~\ref{fig-intro} (a), under the premise of a constant total token number, we investigate the performance of Vid-LLMs by dynamically adjusting the sampling frequency and frame token number.
The results clearly show that both low sampling frequency and low frame token number lead to performance degradation.
Low-frequency sampling results in a sparse image collection as input, omitting crucial temporal details.
On the other hand, excessive frame-feature compression degrades the semantic and spatial contexts of each frame.
When confronted with fine-grained video understanding tasks, this limitation becomes significantly evident, as depicted in Figure~\ref{fig-intro} (b).
Existing models (\eg, LLaMA-VID~\cite{li2023llamavid}) often overlook crucial details early in the input due to low-frequency sampling, leading to inaccuracy.

To address this challenge, we introduce SlowFocus, which is designed to pinpoint relevant temporal segments in response to questions, and subsequently maintains high-quality temporal details to enrich fine-grained video comprehension.
For video understanding tasks, we assume that the relevant details are concentrated within one or several clips.
SlowFocus initially identifies these segments based on the provided questions.
It then densely samples the segmented temporal clips at a high-frequency to extract local temporal features highly pertinent to the questions.
To effectively model the temporal relationships between frames and capture inter-frame contexts, we propose a specialized temporal encoder and a multi-frequency mixing attention module for enhanced temporal comprehension.

To enhance Vid-LLMs' ability to perform SlowFocus based on mixed frequencies, we propose a set of training and inference strategies to improve their temporal localization and fine-grained temporal reasoning capability.
Following VTimeLLM~\cite{huang2023vtimellm}, we fine-tune our Vid-LLM in the second stage on dense video captioning and temporal grounding tasks.
This process enhances the model's ability to predict discrete bins that define the relevant temporal segments.
In the third stage, we adapt the Vid-LLM to the SlowFocus mechanism for high-quality, fine-grained temporal-related tasks, which enables our model to reason precisely based on high-frequency temporal details.


In addition to the scarcity of methods for fine-grained video understanding, existing benchmarks~\cite{xu2016msr, yu2019activitynet} also fall short in providing adequate challenges for specific temporal-related tasks.
To bridge this gap and evaluate our proposed framework, we introduce FineAction-CGR, a newly dedicated benchmark that focuses on fine-grained video understanding, especially reasoning tasks based on temporal details.
Our method demonstrates superior performance on the proposed benchmark, offering a promising solution to high-quality video understanding.


The contributions of this paper are summarized as follows:
\textbf{(i)}
We introduce SlowFocus, a straightforward yet effective framework designed to resolve the prevalent trade-off in existing Vid-LLMs between capturing limited frame-level details and overarching video-level contexts.
SlowFocus adeptly maintains high-frequency local details alongside low-frequency global contexts, facilitating the identification of pertinent temporal segments and precise reasoning on video contents.
\textbf{(ii)}
We present a novel training strategy specifically designed to enhance the temporal localization abilities of Vid-LLMs, and seamlessly adapt them to our newly proposed SlowFocus approach.
\textbf{(iii)}
We establish a comprehensive new benchmark and carry out extensive experiments to rigorously assess the fine-grained video understanding capabilities of Vid-LLMs.
The empirical evidence strongly indicates that our SlowFocus approach significantly outperforms existing models, particularly in tasks requiring detailed temporal understanding and reasoning within videos.

\section{Related works}

\noindent\textbf{Vision large language models.} Researchers have made significant efforts to enable Large Language Models(LLMs) to comprehend visual information.
BLIP-2~\cite{li2023blip} aligns vision-language representation with a lightweight Querying Transformer by concept of Q-Former.
MiniGPT-4~\cite{zhu2023minigpt} aligns detailed image descriptions with advanced LLM which significantly enhances its multi-modal abilities.
Exploring diverse multi-modal instruction-following data, LLaVA~\cite{liu2024visual} has demonstrated impressive multi-model chat abilities.
Recent LLMs, like Kosmos-2~\cite{peng2023kosmos} and VisionLLM~\cite{wang2024visionllm}, probed into more specific aspects of image comprehension, including referring and grounding~\cite{ranasinghe2024learning}, remarkably enhancing the capability to describe intricate image details.

\noindent\textbf{Video large language models.} The exploration of LLMs' potential has been extended from images to videos, which contributes to emergence of Video LLMs. VideoChat~\cite{li2023videochat} combines fine-tuning and LLM-based video agents and is fine-tuned using a specially designed video-centric instruction dataset. Video-ChatGPT~\cite{maaz2023videochatgpt} proposes a novel human assisted and semi-automatic annotation framework for generation high quality instruction data for videos. Video-LLaMA~\cite{zhang2023videollama}, trained on vision-language branch and audio-language branch separately with same visual data and process, has demonstrated impressive abilities in understanding both visual and auditory content in videos. Video-LLaVA~\cite{lin2023videollava} learns united visual representation by alignment before projection and conducts joint training on images and videos simultaneously. With efforts, Video LLMs have exhibited powerful task-handling capabilities in downstream tasks, like text-video retrieval and video captioning. However, these models remain limited in ability to comprehend fine-grained content. To solve this problem, we introduce a model with powerful capability of fine-grained video understanding.

\noindent\textbf{Fine-grained video understanding.} Comprehending videos in fine-grained aspects requires precisely locating and understanding specific events within a video.
It is roughly divided by previous works~\cite{huang2023vtimellm} into temporal grounding~\cite{anne2017localizing, gao2017tall} and dense video captioning~\cite{krishna2017dense, wang2018bidirectional} tasks.
Temporal grounding demands the model to precisely identify start and end timestamps of video segment according to a given text query.
Dense video captioning requires both temporal localization and captioning for all events within an untrimmed video.
Both tasks are constrained by the labor-intensive nature of annotation, resulting in relatively small datasets.
Moreover, fine-grained video understanding demands a deep comprehension of how objects change and interact over time, particularly at a detailed level.
This complexity necessitates temporal reasoning tasks to effectively analyze these dynamics.
However, current research in this field falls short of adequately addressing these requirements.
To solve this problem, our proposed large-scale dataset provides mass temporal annotations and captions in different dimensions, supplying a wealth of information for training in temporal video grounding and reasoning tasks.
\section{Method}
In this section, we begin with a preliminary overview of video LLMs (Vid-LLMs) in Section~\ref{preliminary}.
Following that, we offer a comprehensive explanation of our proposed SlowFocus in Section~\ref{approach}, followed by a detailed introduction to our innovative training strategy in Section~\ref{training}.

\subsection{Preliminary}\label{preliminary}
Contemporary Vid-LLMs typically feature a modular architecture, which includes a visual encoder $E_{V}$, a series of visual adapters $Q$, and a large language model $L$.
For a given video $V = \{V(t) \in \mathbb{R}^{H \times W \times 3} | t=0,...,T \}$ that consists of $T$ frames, along with its associated question $q$, Vid-LLMs generally perform downsampling on the original video $V$ at a fixed interval $M_L$.
This process results in sparsely distributed frames $V_L$:

\begin{equation}
V_L = \{V(M_{L}t) \in \mathbb{R}^{H \times W \times 3} | t=0,...,T \},
\end{equation}

where $M_L \gg 1$.
Consequently, only a very limited number of frames ($T/M_L \ll T$) are selected as the actual input for the Vid-LLM, a technique we refer to as low-frequency sampling in this study.

Subsequently, the visual encoder processes the downsampled frames $V_L$ and encodes them into a series of visual tokens denoted as $z_{L} = E_{V}(V_L)$.
These visual tokens are then transformed to align with the embedding space of the language model through the visual adapter $Q$, resulting in aligned visual tokens $h_L=Q(z_L)$.
Concurrently, the input text query $q$ is encoded into linguistic tokens $h_{q}$ by the textual encoder.
These visual and text tokens are concatenated into a unified sequence $[h_{L}, h_{q}]$, which then serves as the input for the decoder component of the large language model $L$.
The model leverages this combined representation to infer the appropriate answer $ans = L([h_{L}, h_{q}])$, showcasing its ability to perform cross-modal reasoning and respond to human queries.

Although this paradigm is well developed in the field of video understanding, it encounters significant limitations in tasks requiring fine-grained temporal reasoning.
By retaining only a small, discontinuous subset of frames ($t$, where $t \ll T$), substantial information loss can occur, as depicted in Figure~\ref{fig-intro} (b).
Fine-grained temporal video understanding demands that the model concentrate on one or more specific temporal intervals, which could be potentially brief.
However, the low-frequency sampling method predominantly captures overarching information while neglecting crucial local details, leading to inaccuracies such as incorrect responses or hallucinations.
To address these shortcomings, we develop the SlowFocus strategy.

\subsection{SlowFocus}\label{approach}
\begin{figure*}[t]
    \begin{center}
        \includegraphics[width=1.0\linewidth]{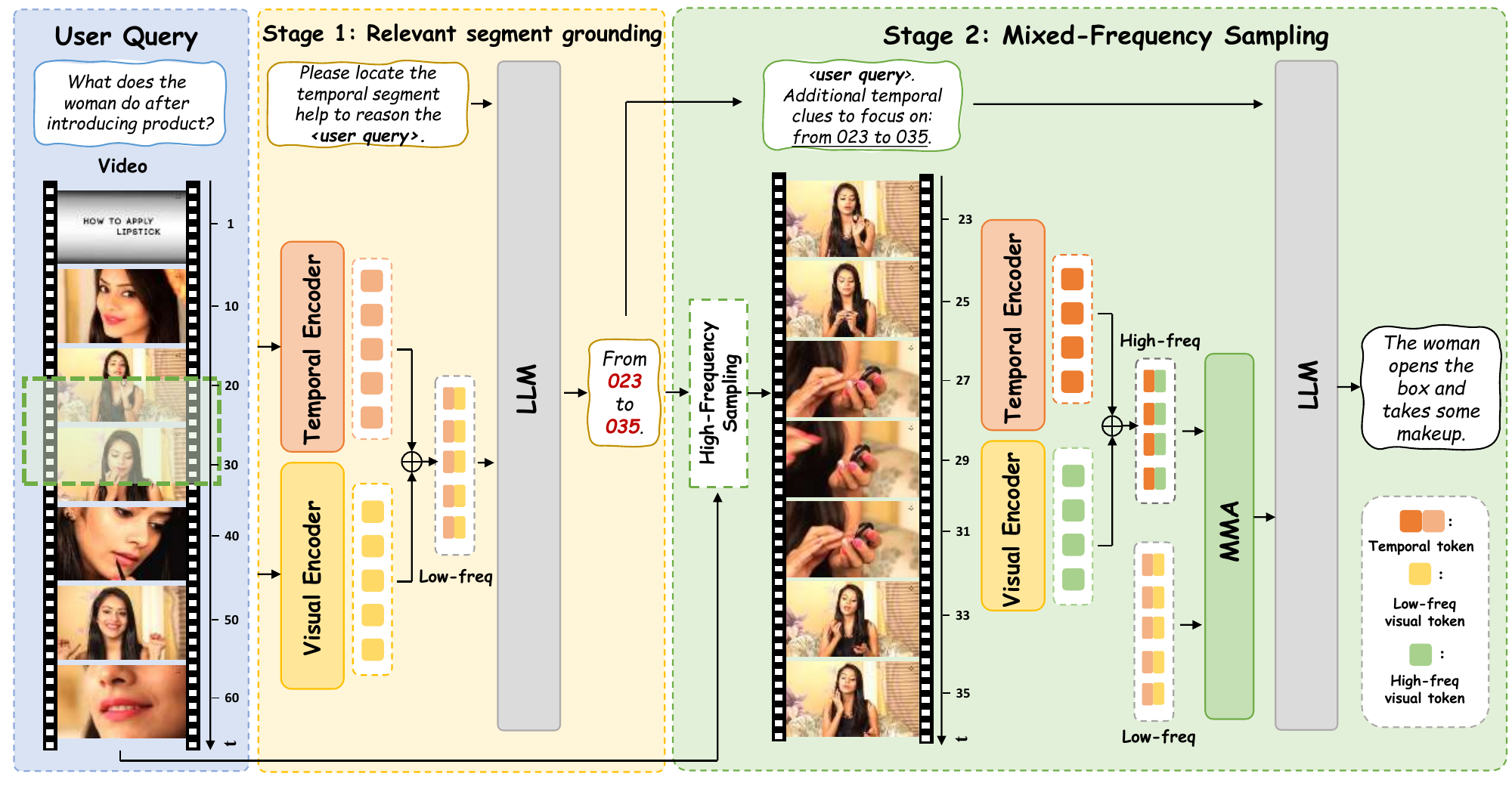}
    \end{center}
    \vspace{-4mm}
    \caption{The framework of SlowFocus. We initially identify the relevant temporal segments based on the given query. Subsequently the high-frequency sampling is performed on these segmented clips. Combined with low-frequency sampling across the entire video, our SlowFocus mechanism maintains mixed-frequency visual tokens to accurately answer the query.}
    \label{fig-method}
    \vspace{-2mm}
\end{figure*}
To improve Vid-LLMs with the SlowFocus mechanism, we first expand the traditional training and inference paradigm by introducing Mixed-Frequency Sampling (MFS).
Subsequently, we explicitly model the temporal relationships among the sampled frames.
Finally, we enhance the visual tokens with our newly proposed multiple-frequency mixing attention, designed to capture long-term contexts.
We now provide a detailed elaboration on each of these components.

\noindent\textbf{Relevant segment grounding.}
To better mimic human cognition in our Vid-LLM, we transform its inference paradigm into a multi-round dialogue format that integrates both the query and temporal awareness elements.
The methodology of this transformation is illustrated in Figure~\ref{fig-method}.
Initially, we sample the entire video at a fixed interval to capture the low-frequency frames $V_L$, which are then encoded into visual embeddings $h_L$.
Following that, we reformat the original question $q$ into temporal grounding questions $q_1$, such as ``\textit{<video>$\backslash$nPlease provide the temporal segment help to reason the question: <question>}'', where \textit{<question>} refers to the original question $q$.

With the augmented question $q_{1}$, we direct the Vid-LLM to identify the relevant temporal segments $\tau$ within the video that are pertinent to the original question $q$.
This is accomplished by providing the model with low-frequency frames containing global information:

\begin{equation}
\tau = L([h_{L}, h_{q_1}]),
\end{equation}

where $h_{q_1}$ represents the textual embedding of question $q_{1}$ and $\tau \subset [0, T]$.
The identified segment $\tau$ is designed to encompass query-related details that are crucial for addressing the question $q$.

\noindent\textbf{Mixed-frequency sampling.}
Subsequently, we perform dense sampling on the identified segment $\tau$ to obtain high-frequency frames $V_H$:

\begin{equation}
V_H = \{V(M_{H}t) \in \mathbb{R}^{H \times W \times 3} | t \in \tau\}, M_H = max(\frac{|\tau|}{N_H}, 1).
\end{equation}
We dynamically adjust the sampling interval $M_{H}$ based on the temporal length of segment $|\tau|$ and to ensure an adequate number of samples $N_H$ are taken from the relevant segment.
The densely sampled frames $V_H$ are then encoded into high-frequency visual tokens $h_H$ by visual encoder.

With the inclusion of local details from high-frequency frames, we augment the initial question $q$ with prompts, such as ``\textit{Additional temporal clues to focus on: ...}'', to form $q_2$, as illustrated in Figure~\ref{fig-method}.
We then combine the mixed-frequency visual tokens to predict the final answer:

\begin{equation}\label{eq-llm}
ans = L([h_L, \pi(h_L, h_H), h_{q_2}]),
\end{equation}
where $\pi$ is the multiple-frequency mixing attention, which will be detailed subsequently.

\begin{figure*}[t]
    \begin{center}
        \includegraphics[width=1.0\linewidth]{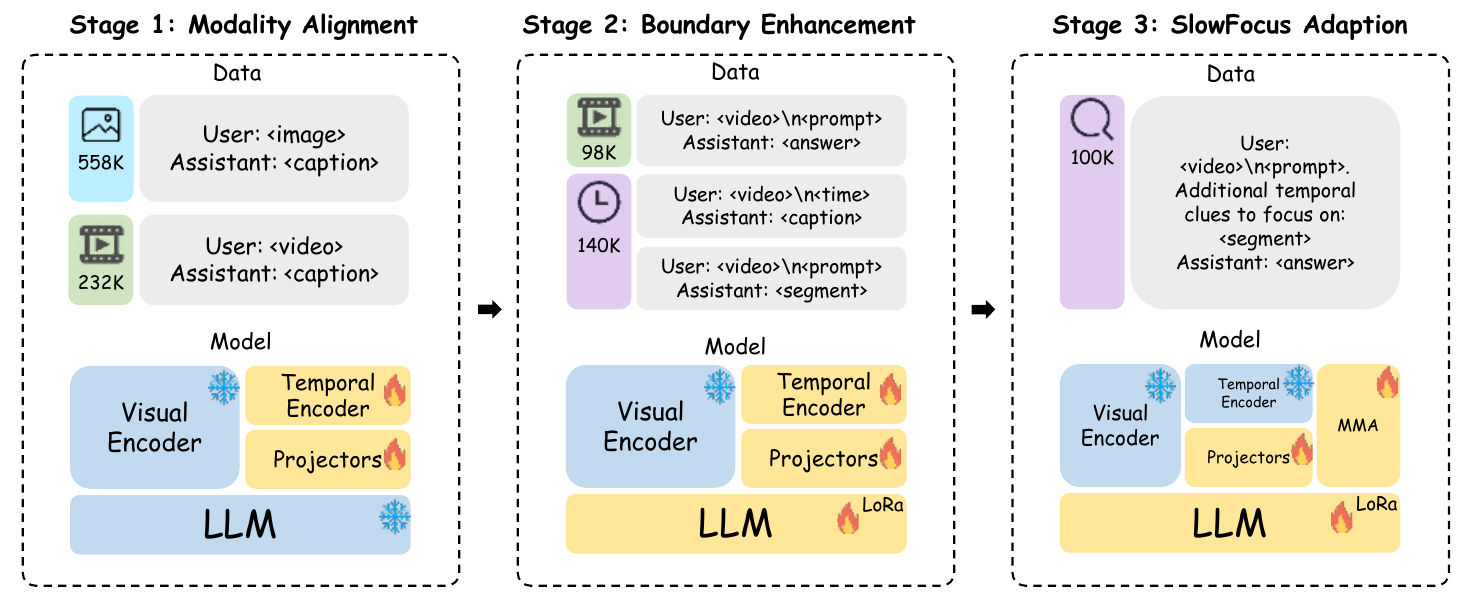}
    \end{center}
    \vspace{-4mm}
    \caption{The training strategy of SlowFocus, including data distribution and parameter updating in each stage. \texttt{<image>} and \texttt{<video>} denote the tokens for image and video, respectively.}
    \label{fig-training}
    \vspace{-2mm}
\end{figure*}

\noindent\textbf{Multiple-frequency mixing attention.}
Because of the roughness of the low-frequency and the detailed specificity of the high-frequency visual features, simply concatenating these two directly into the language model may not yield optimal results.
Fine-grained temporal understanding often requires modeling relationships between multiple events, necessitating the integration of global contexts into high-frequency local features.
From this perspective, we introduce the Multiple-frequency Mixing Attention (MMA):
\begin{equation}
\pi(h_L, h_H) = softmax(\frac{h_{H}h_{L}^{T}}{\sqrt{d}})h_L,
\end{equation}
where, for simplicity, we omit the process of applying FFN to $h_L$ and $h_H$.
The resulting output is then fed into LLM to predict the textual response, as described in Equation~\ref{eq-llm}.

\noindent\textbf{Temporal relationship modeling.}
While LLM can implicitly capture temporal relationships among input visual embeddings based on their sequential positioning, it faces difficulties when the visual tokens are not uniformly distributed along the timeline.
To address this challenge, we propose a temporal encoder that encodes the relative positions of frames into a set of discretized temporal tokens $\mathcal{E} \in \mathbb{R}^{N \times C}$, where $N$ represents the discrete temporal space.
For frames sampled at multiple frequencies $V(t)$, the corresponding temporal token is $\epsilon_{i}$, where $i = \lfloor N * t/T \rfloor$.
These temporal embeddings are then incorporated into the visual tokens by directly adding them to the frame features.

\begin{figure*}[t]
\begin{center}
    \includegraphics[width=1.0\linewidth]{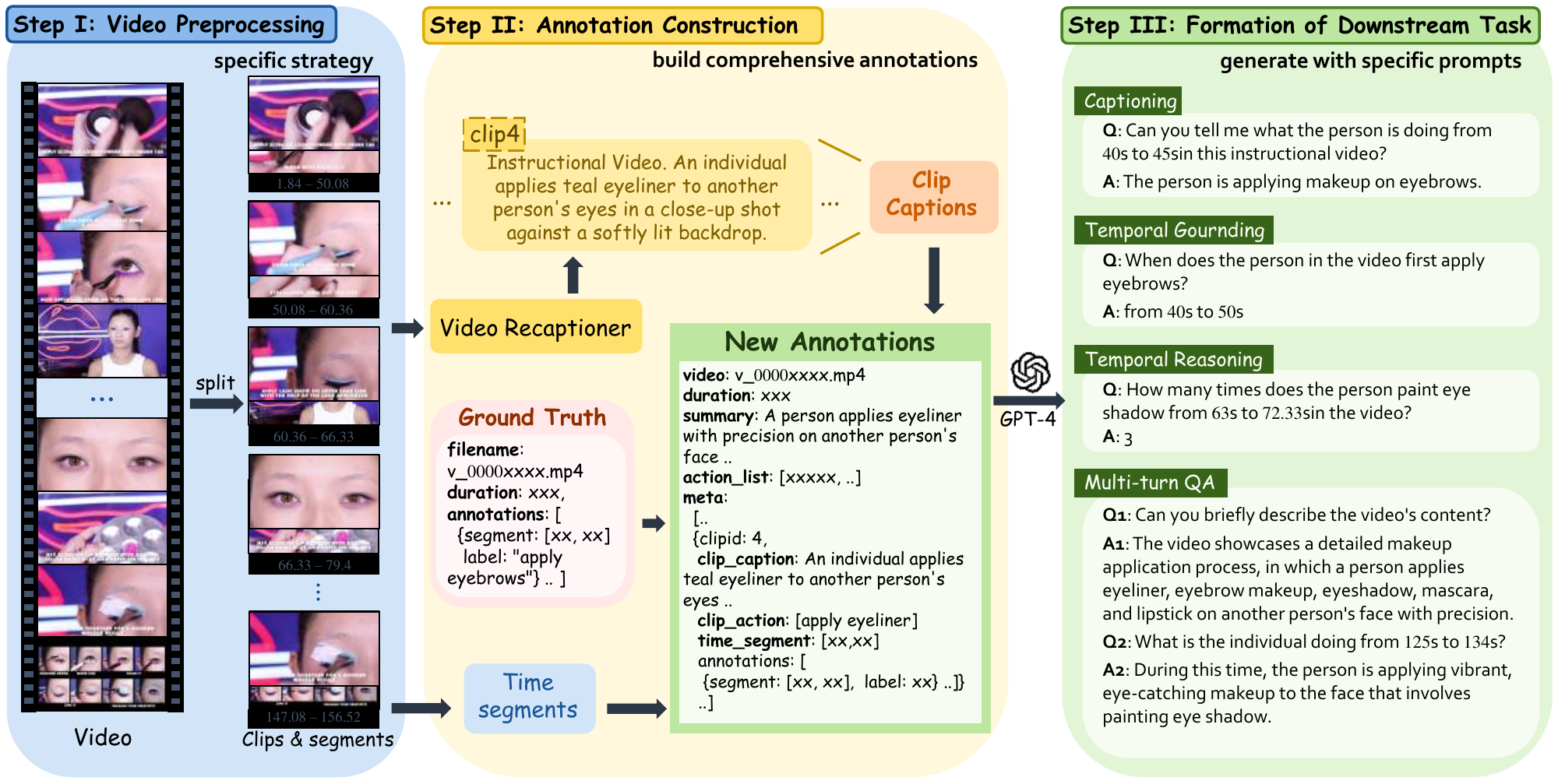}
\end{center}
\vspace{-3mm}
\caption{Pipeline of instruction-following data generation. Split the filtered FineAction videos into clips and extract time segments. Use the fine-tuned Video Recaptioner Model and GPT-4V to generate captions for both video clips and full videos. Integrate the ground truth data, new captions, and time segments to create comprehensive annotations. Finally, generate QA pairs for various tasks using different prompts via GPT-4.}
\label{imgbenchmark}
\vspace{-3mm}
\end{figure*}

\subsection{Training strategy}\label{training}
Considering training efficiency, we introduce a novel training procedure which is structured into three distinct phases in this work: (i) pre-training for modality alignment, (ii) fine-tuning for enhancing temporal grounding, and (iii) SlowFocus adaption, as illustrated in Figure~\ref{fig-training}.
We now provide detailed elaborations on the training strategies and datasets employed for each of these stages. 

\noindent\textbf{Modality alignment.}
In the pre-training stage, our primary focus is to optimize the visual adapters and temporal encoders, while the visual encoder and language model are frozen to ensure that the visual features align effectively with the language space. Following the approach used in LLaMA-VID~\cite{li2023llama}, we utilize the image-text LCS-558K dataset from LLaVA~\cite{liu2024visual}, along with 232K video-caption samples from the WebVid 2.5M~\cite{bain2021frozen}.

\noindent\textbf{Boundary enhancement.}
After the pre-training stage, the Vid-LLM gains proficiency in processing visual information.
During the fine-tuning stage, we focus on enhancing the model's ability to comprehend sequences of video frames, thereby improving its temporal localization capabilities.
In alignment with practices from VTimeLLM~\cite{huang2023vtimellm}, we employ the InternVid-10M-FLT dataset~\cite{wang2023internvid}, which is specifically designed for temporal-awareness training.
The tasks within this dataset include: (i) dense video captioning, which requires detailed descriptions of all events along with their corresponding timestamps, and (ii) segment captioning and temporal video grounding, which involve generating descriptions based on timestamps or determining timestamps based on descriptions.
Throughout this stage, we continue to train the visual adapters and temporal encoders.
Additionally, the LLM is further trained using LoRA~\cite{hu2021lora} to refine its capabilities.

\noindent\textbf{SlowFocus adaption.}
Following the fine-tuning stage, our model has demonstrated the ability to comprehensively understand activities within the video and accurately align them with their respective timestamps.
In the final stage, to further enhance multi-modality comprehension and integration with the MMF mechanism, we construct an instruction-tuning dataset using samples from ActivityNet Captions~\cite{krishna2017dense} and FineAction~\cite{liu2022fineaction}.
We transform the annotations into over 100K high-quality QA dialogues, which will be detailed in the subsequent section.
In alignment with our SlowFocus mechanism, during training with fine-grained captioning and reasoning tasks, the model is provided with the ground-truth temporal segments $\tau_{GT}$ and subjected to mixed-frequency sampling.
In this stage, we freeze all the visual adapters and attention modules, only training the language model as well as the MMA module.
The parameters of visual encoder are frozen all over the stages.
\section{FineAction-CGR benchmark}
In addition to the scarcity of previous methods for fine-grained video understanding, existing benchmarks~\cite{xu2016msr, yu2019activitynet} do not adequately challenge specific temporal-related tasks.
To address this gap and evaluate our proposed framework, we introduce FineAction-CGR, a comprehensive and high-quality instruction-following benchmark designed for evaluating the capabilities of Video LLMs in fine-grained video understanding.
FineAction~\cite{liu2022fineaction} is a large-scale and fine-grained video dataset encompassing diverse action categories with detailed annotations of action instances and time segments.
For the purpose of SlowFocus adaption and model evluation, we divide the FineAction dataset into training and testing sets based on videos, allocating 75\% to the training set and 25\% to the test set, ensuring these is no overlap between them.
Building on this foundation, we probe time information and content information from FineAction and expand it into multi-tasks.
This section outlines the data construction procedures, which consist of three main steps: (i) video preprocessing, (ii) annotation construction, and (iii) formation of downstream tasks.
More details are included in the appendix.

\begin{table}[t]
\renewcommand\arraystretch{1.5}
\begin{center}
\resizebox{1.0\linewidth}{!}{
\begin{tabular}{ccc|cccc|cccc|cc}
\midrule[1.0pt]
\rowcolor[gray]{0.8}& & & \multicolumn{4}{c|}{Temporal grounding} & \multicolumn{4}{c|}{Temporal captioning} & \multicolumn{2}{c}{Temporal reasoning} \\
\rowcolor[gray]{0.8}\multirow{-2}{*}{Method} & \multirow{-2}{*}{LLM} & \multirow{-2}{*}{LoRA} & mIoU & R@0.3 & R@0.5 & R@0.7 & B & M & R & C & Acc & Score \\
\midrule[1.0pt]
VideoLLaMA~\cite{zhang2023videollama} & Vicuna-7B & \xmark & 0.17 & 0.25 & 0.11 & 0.00 & 0.06 & 0.09 & 0.08 & 0.13 & 8.75 & 0.53 \\
Video-ChatGPT~\cite{maaz2023videochatgpt} & Vicuna-7B & \xmark & 0.06 & 0.10 & 0.01 & 0.00 & 0.15 & 0.12 & 0.10 & 0.21 & 13.93 & 0.79 \\
LLaMA-VID~\cite{li2023llama} & Vicuna-7B & \xmark & 0.35 & 0.52 & 0.17 & 0.03 & 0.16 & 0.12 & 0.11 & 0.23 & 15.65 & 0.87 \\
VTimeLLM~\cite{huang2023vtimellm} & Vicuna-7B & \cmark & 27.69 & 32.83 & 24.26 & 21.87 & 0.05 & 0.09 & 0.08 & 0.12 & 9.96 & 0.54 \\
\midrule
LLaMA-VID$\dagger$ & Vicuna-7B & \cmark & 22.38 & 26.17 & 19.47 & 17.53 & 0.23 & 0.20 & 0.37 & 1.03 & 24.81 & 1.26 \\
Ours & Vicuna-7B & \cmark & \textbf{66.68} & \textbf{85.80} & \textbf{73.01} & \textbf{56.25} & \textbf{0.66} & \textbf{0.41} & \textbf{0.70} & \textbf{3.27} & \textbf{53.10} & \textbf{2.78} \\
\midrule[1.0pt]
\end{tabular}}
\end{center}
\caption{Main results on FineAction-CGR benchmark. The column \textit{LoRA} represents whether the LLM is fine-tuned fully or using LoRA. $\dagger$: Model is re-trained on the stage 3's data. B: B@4. M: METEOR. R: ROUGE. C: CIDEr.}
\label{tab-main}
\vspace{-6mm}
\end{table}
\begin{table}[t]
\renewcommand\arraystretch{1.5}
\begin{center}
\resizebox{1.0\linewidth}{!}{
\begin{tabular}{ccc|cc|cc|cc|ccccc}
\midrule[1.0pt]
\rowcolor[gray]{0.8}& & & \multicolumn{2}{c|}{MSVD-QA} & \multicolumn{2}{c|}{MSRVTT-QA} & \multicolumn{2}{c|}{ActivityNet-QA} & \multicolumn{5}{c}{Video-based generative performance} \\
\rowcolor[gray]{0.8}\multirow{-2}{*}{Method} & \multirow{-2}{*}{LLM} & \multirow{-2}{*}{LoRA} & Acc & Score & Acc & Score & Acc & Score & Correctness & Detail & Context & Temporal & Consistency \\
\midrule[1.0pt]
FrozenBiLM~\cite{yang2022zero} & DeBERTa-V2 & \xmark & 32.2 & - & 16.8 & - & 24.7 & - & - & - & - & - & - \\
VideoLLaMA~\cite{zhang2023videollama} & Vicuna-7B & \xmark & 51.6 & 2.5 & 29.6 & 1.8 & 12.4 & 1.1 & 1.96 & 2.18 & 2.16 & 1.82 & 1.79 \\
LLaMA-Adapter~\cite{zhang2023llama} & LLaMA-7B & \xmark & 54.9 & 3.1 & 43.8 & 2.7 & 34.2 & 2.7 & 2.03 & 2.32 & 2.30 & 1.98 & 2.15 \\
VideoChat~\cite{li2023videochat} & Vicuna-7B & \xmark & 56.3 & 2.8 & 45.0 & 2.5 & 26.5 & 2.2 & 2.23 & 2.50 & 2.53 & 1.94 & 2.24 \\
Video-ChatGPT~\cite{maaz2023videochatgpt} & Vicuna-7B & \xmark & 64.9 & 3.3 & 49.3 & 2.8 & 35.2 & 2.7 & 2.40 & 2.52 & 2.62 & 1.98 & 2.37 \\
LLaMA-VID~\cite{li2023llama} & Vicuna-7B & \xmark & 69.7 & 3.7 & 57.7 & 3.2 & 47.4 & 3.3 & \textbf{2.96} & 3.00 & 3.53 & 2.46 & 2.51 \\
\midrule
LLaMA-VID & Vicuna-7B & \cmark & 69.2 & 3.4 & 57.1 & 2.9 & 45.6 & 3.3 & 2.87 & 2.89 & 3.28 & 2.13 & 2.47 \\
Ours & Vicuna-7B & \cmark & \textbf{70.1} & \textbf{3.9} & \textbf{58.3} & \textbf{3.5} & \textbf{48.4} & \textbf{3.6} & 2.95 & \textbf{3.03} & \textbf{3.61} & \textbf{2.54} & \textbf{2.60} \\
\midrule[1.0pt]
\end{tabular}}
\end{center}
\caption{Comparison with existing methods on coarse-grained video understanding benchmarks. Our method achieve on par performance with state-of-the-art models.}
\label{tab-coarse}
\vspace{-8mm}
\end{table}

\textbf{Video preprocessing.}
As depicted in Step \uppercase\expandafter{\romannumeral1} of Figure~\ref{imgbenchmark}, we employ a modified two-stage splitting algorithm from Panda70M~\cite{chen2024panda} to segment videos into clips, thereby providing raw materials for fine-grained information at the clip level.
This preprocessing step yields 62,912 video clips.

\textbf{Annotation construction.} Step \uppercase\expandafter{\romannumeral2} in Figure~\ref{imgbenchmark} outlines our approach to creating new annotations, which include time segments, captions, and action labels.
Given the high cost of using GPT-4V~\cite{achiam2023gpt} for generating large-scale, clip-level captions, we generate detailed captions for entire videos with GPT-4V and then use a fine-tuned Video Recaptioner Model to produce large-scale, clip-level captions.
By integrating action labels and time segments from the ground truth with clip segments and the generated captions, we create comprehensive new annotations.

\textbf{Formation of downstream tasks.} Utilizing the new annotations, we design four types of video instruction-following tasks, as illustrated in Step \uppercase\expandafter{\romannumeral3} of Figure~\ref{imgbenchmark}: (i) segmented captioning, (ii) temporal video grounding, (iii) temporal video reasoning and (iv) multi-turn QA.
Detailed information on construction procedures, task formats, prompts, and case studies can be found in Appendix~\ref{app:b}-~\ref{app:d}.

\begin{minipage}[!t]{\textwidth}
 \begin{minipage}[t]{0.62\textwidth}
  \centering
  \makeatletter\def\@captype{table}\makeatother
  \resizebox{1.0\textwidth}{0.06\textheight}{
   \begin{tabular}{ccc|cc|cc}
    \midrule[1.0pt]
    \rowcolor[gray]{0.8}& & & \multicolumn{2}{c|}{Global mode} & \multicolumn{2}{c}{Breakpoint mode} \\
    \rowcolor[gray]{0.8}\multirow{-2}{*}{Method} & \multirow{-2}{*}{LLM} & \multirow{-2}{*}{LoRA} & Acc & Score & Acc & Score \\
    \midrule[1.0pt]
    VideoChat~\cite{li2023videochat} & Vicuna-7B & \xmark & 57.8 & 3.00 & 46.1 & 2.29 \\
    VideoLLaMA~\cite{zhang2023videollama} & Vicuna-7B & \xmark & 51.7 & 2.67 & 39.1 & 2.04\\
    Video-ChatGPT~\cite{maaz2023videochatgpt} & Vicuna-7B & \xmark & 47.6 & 2.55 & 48.0 & 2.45 \\
    MovieChat~\cite{song2024moviechat} & Vicuna-7B & \xmark & 62.3 & 3.23 & 48.3 & 2.57 \\
    \midrule
    Ours & Vicuna-7B & \cmark & 58.6 & 3.14 & 48.1 & 2.53\\
    \midrule[1.0pt]
    \end{tabular}}
   \caption{Comparison with existing methods on MovieChat-1K.}
   \label{tab-moviechat}
 \end{minipage}
 \begin{minipage}[t]{0.37\textwidth}
  \centering
  \makeatletter\def\@captype{table}\makeatother
  \resizebox{1.0\textwidth}{0.06\textheight}{
   \begin{tabular}{ccc|c}
    \midrule[1.0pt]
    \rowcolor[gray]{0.8}Method & LLM & LoRA & Acc \\
    \midrule[1.0pt]
    FrozenBiLM~\cite{frozenbilm} & DeBERTa~\cite{he2020deberta} & \xmark & 26.9 \\
    VIOLET~\cite{fu2021violet} & - & \xmark & 19.9 \\
    InternVideo~\cite{wang2022internvideo} & - & \xmark & 32.1 \\
    LLoVi-7B~\cite{zhang2023llovi} & Llama2-7B~\cite{touvron2023llama2} & - & 34.0 \\
    Vamos~\cite{wang2023vamos} & GPT-4 & - & 36.7 \\
    LangRepo-12B~\cite{kahatapitiya2024langrepo} & Mixtral~\cite{jiang2024mixtral} & - & 41.2 \\
    \midrule
    Ours & Vicuna-7B & \cmark & 39.7 \\
    \midrule[1.0pt]
    \end{tabular}}
    \caption{Comparison with existing methods on EgoSchema.}
    \label{tab-egoschema}
 \end{minipage}
\vspace{-1mm}
\end{minipage}
\begin{table}[h]
\renewcommand\arraystretch{1.5}
\begin{center}
\resizebox{1.0\linewidth}{!}{
\begin{tabular}{ccc|cccc|cc}
\midrule[1.0pt]
\rowcolor[gray]{0.8}& & & \multicolumn{4}{c|}{Temporal grounding} & \multicolumn{2}{c}{Temporal reasoning} \\
\rowcolor[gray]{0.8}\multirow{-2}{*}{Sampling strategy} & \multirow{-2}{*}{Temporal encoder} & \multirow{-2}{*}{MMA} & mIoU & R@0.3 & R@0.5 & R@0.7 & Acc & Score \\
\midrule[1.0pt]
$V_L$ & \xmark & \xmark & 32.54 & 41.67 & 34.52 & 25.84 & 30.25 & 1.57 \\
$V_L + V_H, N_H=10$ & \xmark & \xmark & 34.81 & 44.65 & 36.49 & 27.90 & 39.12 & 2.02 \\
$V_L + V_H, N_H=10$ & \cmark & \xmark & 57.26 & 73.47 & 62.28 & 47.16 & 46.37 & 2.37 \\
$V_L + V_H, N_H=20$ & \cmark & \xmark & 64.92 & 83.25 & 70.19 & 55.13 & 51.68 & 2.66 \\
$V_L + V_H, N_H=40$ & \cmark & \xmark & 65.03 & 82.96 & 70.23 & 56.01 & 51.59 & 2.62 \\
\rowcolor[gray]{0.9}$V_L + V_H, N_H=20$ & \cmark & \cmark & 66.68 & 85.80 & 73.01 & 56.25 & 53.10 & 2.78 \\
\midrule[1.0pt]
\end{tabular}}
\end{center}
\caption{Components analysis. $V_L$ means only low-frequency frames are sampled. $V_L+V_H$ represents performing mixed-frequency sampling and $N_H$ denotes the number of high-frequency frames.}
\label{tab-component}
\vspace{-4mm}
\end{table}

\textbf{Evaluation protocol.}
To assess the capability of video language model in comprehending fine-grained video understanding tasks, we integrate multiple evaluation metrics in comprehensively evaluate the model's performance against our proposed benchmark.
For temporal grounding task, we calculate the Intersection over Union (IoU) between the temporal segments generated by the model and the corresponding ground truth, and we report meanIoU (mIoU) metric.
For segmented captioning, we employ commonly accepted captioning-based metric, including BLEU~\cite{papineni2002bleu}, METEOR~\cite{banerjee2005meteor}, ROUGE~\cite{lin2004rouge} and CIDEr~\cite{vedantam2015cider}.
For temporal reasoning task, we utilize GPT-4~\cite{achiam2023gpt} to evaluate the accuracy and score of the generated answers.

\section{Experiments}
\label{sec:exp}
\subsection{Experiment setup}
In our experiments, we implement LLaMA-VID~\cite{li2023llamavid} as baseline and utilize Vicuna-7B v1.5~\cite{zheng2023judging} as our foundational LLM.
More implementation details are provided in the appendix.
We adjust the resolution of input videos to 224 $\times$ 224, and each frame is condensed into 64 tokens.
The low-frequency sampling interval $M_L$ is set to match the original video fps, ensuring one frame is sampled every second.
We define the dense sampling number $N_H$ as 20 and the size of temporal token space $N$ as 1000.
The AdamW~\cite{loshchilov2017decoupled} optimizer is applied with a cosine learning rate and decay and a warm-up period.
We train our Vid-LLM for three stages.
During the initial pre-training stage, the learning rate is set to $1 \times 10^{-3}$.
For the subsequent fine-tuning stages, the learning rate is adjusted to $2 \times 10^{-4}$.
Additionally, the LoRA parameters are configured with $r = 64$ and $alpha = 128$.
All experiments are conducted on 8 V100 GPUs.



\subsection{Main results}
\noindent\textbf{Results on fine-grained video understanding.}
We evaluate the performance of existing Vid-LLMs on our proposed fine-grained video understanding benchmark FineAction-CGR, as detailed in Table~\ref{tab-main}.
Further specifics are available in the appendix.
Our method significantly outperforms other counterparts, achieving a mIoU of 66.68 for temporal grounding and an accuracy of 53.10\% for the reasoning task.
Notably, most other models, with the exception of VTimeLLM~\cite{huang2023vtimellm}, exhibit subpar performance in both the temporal grounding and reasoning tasks on the benchmark.
We propose several possible explanations for this observation:
(i)
These models lack sensitivity to precise time boundaries, making it challenging for them to accurately predict temporal segments.
(ii)
Due to their reliance on low-frequency sampling, these models struggle to capture fine-grained temporal details, which adversely affects their performance on tasks requiring fine-grained temporal reasoning.

We also fine-tune the baseline using stage 3.
To ensure fairness, the implementation details for baseline fine-tuning are kept consistent with those of our method.
The baseline's performance improves because stage 3 includes tasks focused on temporal grounding and is specifically designed for fine-grained analysis.
Furthermore, the remaining performance gap further supports our explanations.

\begin{minipage}[!t]{\textwidth}
 \begin{minipage}[t]{0.5\textwidth}
  \centering
  \makeatletter\def\@captype{table}\makeatother
  \resizebox{1.0\textwidth}{0.05\textheight}{
   \begin{tabular}{c|cccc|cc}
    \midrule[1.0pt]
    \rowcolor[gray]{0.8}& \multicolumn{4}{c|}{Temporal grounding} & \multicolumn{2}{c}{Temporal reasoning} \\
    \rowcolor[gray]{0.8}\multirow{-2}{*}{Stages} & mIoU & R@0.3 & R@0.5 & R@0.7 & Acc & Score \\
    \midrule[1.0pt]
    1 & 0.11 & 0.29 & 0.00 & 0.00 & 7.13 & 0.51 \\
    1+2 & 51.67 & 63.29 & 57.84 & 45.81 & 20.34 & 1.04 \\
    1+3 & 28.58 & 36.72 & 33.25 & 24.22 & 32.27 & 1.63 \\
    \rowcolor[gray]{0.9}1+2+3 & 66.68 & 85.80 & 73.01 & 56.25 & 53.10 & 2.78 \\
    \midrule[1.0pt]
    \end{tabular}}
   \caption{Ablation of training stages. The first stage itself yields poor result. Integrating these stages together results in the optimal performance.}
   \label{tab-finetune}
 \end{minipage}
 \begin{minipage}[t]{0.5\textwidth}
  \centering
  \makeatletter\def\@captype{table}\makeatother
  \resizebox{1.0\textwidth}{0.05\textheight}{
   \begin{tabular}{cc|cccc|cc}
    \midrule[1.0pt]
    \rowcolor[gray]{0.8}& & \multicolumn{4}{c|}{Temporal grounding} & \multicolumn{2}{c}{Temporal reasoning} \\
    \rowcolor[gray]{0.8}\multirow{-2}{*}{Fps} & \multirow{-2}{*}{$N_V$} & mIoU & R@0.3 & R@0.5 & R@0.7 & Acc & Score \\
    \midrule[1.0pt]
    64 & 1 & 24.15 & 33.05 & 24.84 & 25.71 & 34.58 & 1.80 \\
    16 & 4 & 45.19 & 58.67 & 50.17 & 38.18 & 45.36 & 2.35 \\
    4 & 16 & 61.01 & 76.54 & 65.12 & 52.63 & 52.17 & 2.67 \\
    \rowcolor[gray]{0.9}1 & 64 & 66.68 & 85.80 & 73.01 & 56.25 & 53.10 & 2.78 \\
    \midrule[1.0pt]
    \end{tabular}}
    \caption{The results of dynamically adjusting sampling frequency (fps) and frame token number $N_V$ when maintaining constant total token number.}
    \label{tab-freq}
 \end{minipage}
\end{minipage}

\noindent\textbf{Results on coarse-grained video-based benchmarks.}
In Table~\ref{tab-coarse}, we present a comparative evaluation of our method against various state-of-the-art methods across three zero-shot video-QA benchmarks: MSVD-QA~\cite{xu2017video}, MSRVTT-QA~\cite{xu2016msr}, and ActivityNet-QA~\cite{yu2019activitynet}.
We also conduct experiments on the video-based generative performance benchmark~\cite{maaz2023videochatgpt}.
The results demonstrate that the proposed SlowFocus mechanism not only enhances fine-grained video understanding but also delivers competitive performance in coarse-grained video understanding tasks, achieving results on par with state-of-the-art models.

\noindent\textbf{Results on long video benchmarks.}
To further investigate the effectiveness of the proposed method on more challenging scenarios, we provide evaluations on long video benchmarks, including MovieChat-1K~\cite{song2024moviechat} and EgoSchema~\cite{mangalam2023egoschema}.
As shown in Table~\ref{tab-moviechat}, we evaluate our method on MovieChat-1K.
The results show that although our method is not specifically trained on long video benchmarks (in contrast, MovieChat~\cite{song2024moviechat} has undergone targeted training for long videos), it still achieves competitive results (58.6\% accuracy in global mode and 48.1\% in breakpoint mode).

Additionally, we also conduct experiments on EgoSchema~\cite{mangalam2023egoschema} benchmark, as detailed in Table~\ref{tab-egoschema}.
The results further demonstrate that, despite not being specifically trained on long video datasets, our method still achieves competitive performance.


\begin{figure*}[t]
    \begin{center}
        \includegraphics[width=1.0\linewidth]{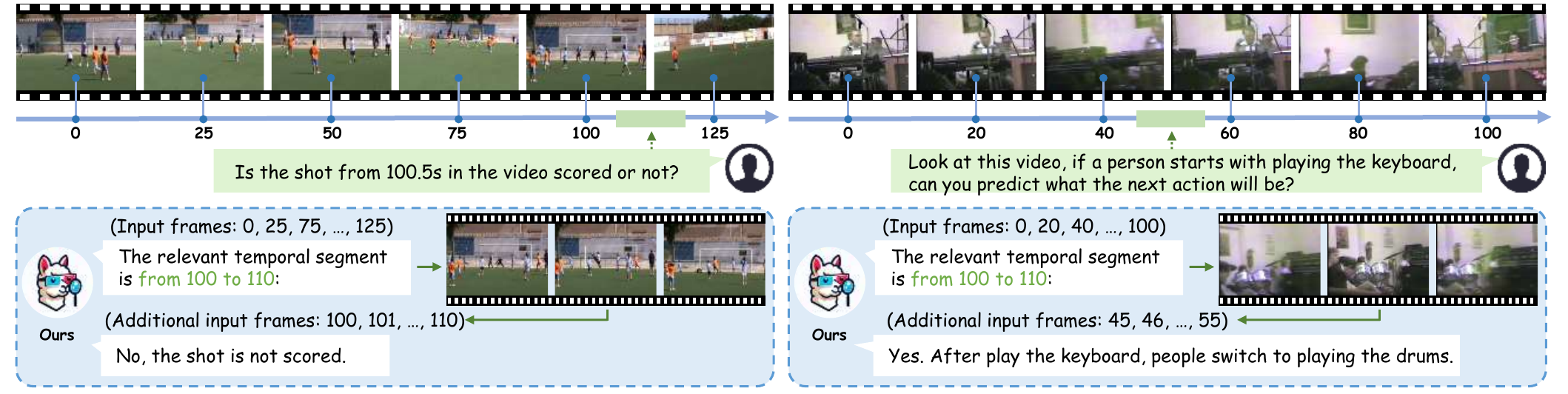}
    \end{center}
    \vspace{-4mm}
    \caption{Qualitative examples. Our proposed SlowFocus can effectively leverages the segmented temporal clues to accurately answer the posed question.}
    \label{fig-vis}
    \vspace{-3mm}
\end{figure*}

\noindent\textbf{Qualitative results.}
Figure~\ref{fig-vis} illustrates the qualitative results of our method on different videos.
Our SlowFocus comprehensively analyzes the entire video, accurately identifies relevant temporal details based on the posed question, and provides precise answer within the context of videos.

\subsection{Ablations}
In this section, we provide detailed ablation studies of our method conducted utilizing Vicuna-7B as the foundation model.

\noindent\textbf{Component-wise analysis.}
We first investigate the impact of each proposed component in Table~\ref{tab-component}.
Importantly, we observe that the baseline model, which relies solely on low-frequency frames ($V_L$), achieves limited performance, with an mIoU of 32.54 and an accuracy of 30.25.
This highlights the challenges faced by current Vid-LLMs in addressing fine-grained temporal tasks using only low-frequency sampling.
Additionally, the results in the second row indicate that the inclusion of MFS significantly improves reasoning capabilities, resulting in an accuracy increase of +8.87.
Furthermore, the temporal encoder, which models temporal relationships effectively, boosts both grounding and reasoning capabilities, with an increase of +22.45 in mIoU and +7.25 in accuracy.
Increasing the high-frequency sampling number $N_H$ from 10 to 20 leads to a noticeable performance gain (+7.66 in mIoU and +5.31 in accuracy), although further increasing $N_H$ to 40 offers only marginal benefits.
Lastly, the MMA module contribute an accuracy enhancement of +1.42.

\begin{table}[t]
\renewcommand\arraystretch{1.5}
\begin{center}
\resizebox{1.0\linewidth}{!}{
\begin{tabular}{c|cccc|cccc|cc}
\midrule[1.0pt]
\rowcolor[gray]{0.8}& \multicolumn{4}{c|}{Temporal grounding} & \multicolumn{4}{c|}{Temporal captioning} & \multicolumn{2}{c}{Temporal reasoning} \\
\rowcolor[gray]{0.8}\multirow{-2}{*}{N} & mIoU & R@0.3 & R@0.5 & R@0.7 & B & M & R & C & Acc & Score \\
\midrule[1.0pt]
0.1K & 43.81 & 62.26 & 51.64 & 33.18 & 0.47 & 0.33 & 0.49 & 2.15 & 39.41 & 2.20 \\
\rowcolor[gray]{0.9}1K & 66.68 & 85.80 & 73.01 & 56.25 & 0.66 & 0.41 & 0.70 & 3.27 & 53.10 & 2.78 \\
10K & 63.74 & 86.19 & 71.05 & 55.17 & 0.66 & 0.40 & 0.71 & 3.24 & 52.59 & 2.71 \\
\midrule[1.0pt]
\end{tabular}}
\end{center}
\caption{Ablation study on temporal token space $N$.}
\label{tab-bins}
\vspace{-2mm}
\end{table}

\noindent\textbf{Necessity of fine-tuning strategy.}
We also analyze the impact of each training stage, as reported in Table~\ref{tab-finetune}.
Directly evaluating the pre-trained model after stage 1 yields poor performance.
Based on necessary pre-training stage 1, when only utilizing stage 2 for boundary enhancement, there is a significant improvement by 51.56 in mIoU in temporal grounding ability.
Furthermore, implementing only stage 3 for SlowFocus adaptation boosts the performance in temporal reasoning by 25.14 in accuracy.
Integrating these stages results in the highest performance in both mIoU and accuracy metrics, highlighting the irreplaceability of our proposed comprehensive training strategy.


\noindent\textbf{What contributes more to fine-grained video understanding?}
In Table~\ref{tab-freq}, we conduct further experiments to explore how changes in sampling frequency and the number of frame tokens influence our method.
For meaningful comparison, we maintain constant total token numbers while gradually decreasing the sampling frequency.
The results indicate that within proposed SlowFocus, performance improves with an increase in frame tokens and remains relatively unaffected by a decrease in global sampling frequency.
This demonstrates the efficacy of our approach in high-frequency sampling and effectively alleviates the trade-off dilemma between sampling frequency and frame tokens.

\noindent\textbf{Ablation on discretized temporal space.}
We investigate the influence of temporal token space $N$ in Table~\ref{tab-bins}.
The results demonstrate that when the token space increase from 0.1K to 1K, a significant improvement (+22.87 in mIoU and +13.69 in accuracy) occurs.
While further increasing $N$ from 1K to 10K, the performance drops instead.

\section{Conclusion and limitations}
\label{sec:conclusion}
\noindent\textbf{Conclusion.}
We introduce SlowFocus, a straightforward yet potent mechanism that significantly enhances fine-grained video understanding.
Our approach improves the temporal localization capabilities of Vid-LLMs, enabling them to precisely identify relevant temporal segments based on the given query.
Moreover, with our newly developed temporal encoder and multi-frequency mixing attention module, our method effectively models the temporal relationships among frames and captures inter-frame context.
Demonstrating superior performance on our newly established fine-grained video understanding benchmarks, we hope that our work can propel the development of Vid-LLMs in achieving advanced video understanding.

\noindent\textbf{Limitations.}
Despite significant advancements made in this work to enhance Vid-LLM's access to high-frequency temporal details and its temporal reasoning capabilities, challenges persist due to the limited existing research on maintaining high resolution in video.
Consequently, our method may still encounter inaccurate predictions stemming from the ambiguity of spatial details.

\section*{Acknowledgements}
This work was supported in part by National Natural Science Foundation of China (Grant No. 62106050 and 62376060) and
Natural Science Foundation of Shanghai (Grant No. 22ZR1407500).

\bibliographystyle{plain}
\bibliography{egbib}
\newpage
\appendix
\section{Overview}
In this appendix we present:
\begin{itemize}
    \item More implementation details (Section~\ref{app:imple}).
    \item Data construction details (Section \ref{app:b}).
    \item Data statistics (Section \ref{app:c}).
    \item Prompts used for instruction generation (Section \ref{app:d}).
    \item Discussion on broader impacts (Section~\ref{appendix:impacts}).
\end{itemize}

\section{More implementation details}\label{app:imple}
In our work, we employ CLIP-ViT-L-14 for the frozen visual encoder, which accepts image resolutions at 224 $\times$ 224 as input.
For visual adapters, we follow LLaVA~\cite{liu2024visual} to train a projector layer, aligning visual features with the pre-trained LLM word embedding.

We also elaborate on how we enable LLM to predict temporal segments during stages 2 and 3 fine-tuning.
Following VTimeLLM~\cite{huang2023vtimellm}, we utilize the textual format \textit{``from s to e''} to denote a video clip, where \textit{s} and \textit{e} represent the starting and ending points, respectively.
These time points are normalized and range from 000 to 999, corresponding to the discretized temporal token space $N$.
During training, Vid-LLM is exposed to supervisory signals as described above, which enable it to develop the capability to accurately locate temporal segments.

\section{Data construction}\label{app:b}
\textbf{Step I: video preprocessing.} We split videos into video clips to provide raw materials for fine-grained information in video clip level. 

A proper splitting and stitching strategy is necessary. For one thing, the content of each clip should be semantically consistent. Without this consistency, actions within the same video clip obtained by erroneous segmentation can't be inferred reasonably and in chronological order, which negatively impacts the performance of downstream tasks such as action recognition and temporal reasoning. For another, the video clips with insufficient information or that are too short are not suitable for subsequent tasks, such as temporal grounding. Therefore, we adopt a modified two-stage splitting algorithm in Panda70M~\cite{chen2024panda} to obtain semantically consistent video clips.

In first stage, We collect time boundaries of events in the video. Instead of detecting abrupt changes in pixels of adjacent frames, we detect lens switching so that actions in the same video clip can be inferred reasonably. In second stage,  we stitch adjacent events based on time boundaries in first stage. Our stitching strategy could be depicted as merging short video clips and that are semantically similar. To be specific, we merge current event with the previous, if one of the following two conditions is met: 1)the video clip duration is less than five seconds, 2) the feature distance between the start frame of the current event and the end frame of the previous event does not exceed 0.1. Finally, we split videos into clips according to time boundaries from second stage. 

Our adjusted splitting strategy has two advantages: 1) it preserves the information density of each segment, ensuring that subsequently generated captions are reasonable; 2) it maintains logical temporal consistency of actions within each segment, which facilitates the generation of question-answer pairs in downstream tasks. Following the preprocessing step, we obtained 62,912 video clips. 

\textbf{Step II: annotation construction.} As Figure~\ref{imgbenchmark} illustrates, new annotations we construct consist of time segments, captions and actions. (1) Action labels are from FineAction ground truth. (2) Time segments cover clip's time segments from above segmentation step and the action time segments from ground truth. (3) Captions cover video captions and clip captions. We first utilzie GPT-4V to generate detailed captions for entire videos. However, due to the cost of GPT-4V  for generating large-scale, clip-level video captions, we fine tune a Video Recaptioner Model based on the captions generated by GPT-4V, which is then utilized to generate large-scale, clip-level captions. With actions and time segments from ground truth, we integrate clip segments from splitting stage and captions generated by GPT-4V and Video Recaptioner Model to get new, comprehensive annotations.

\begin{figure}[!htp]
    \centering
    \includegraphics[width=1.0\linewidth]{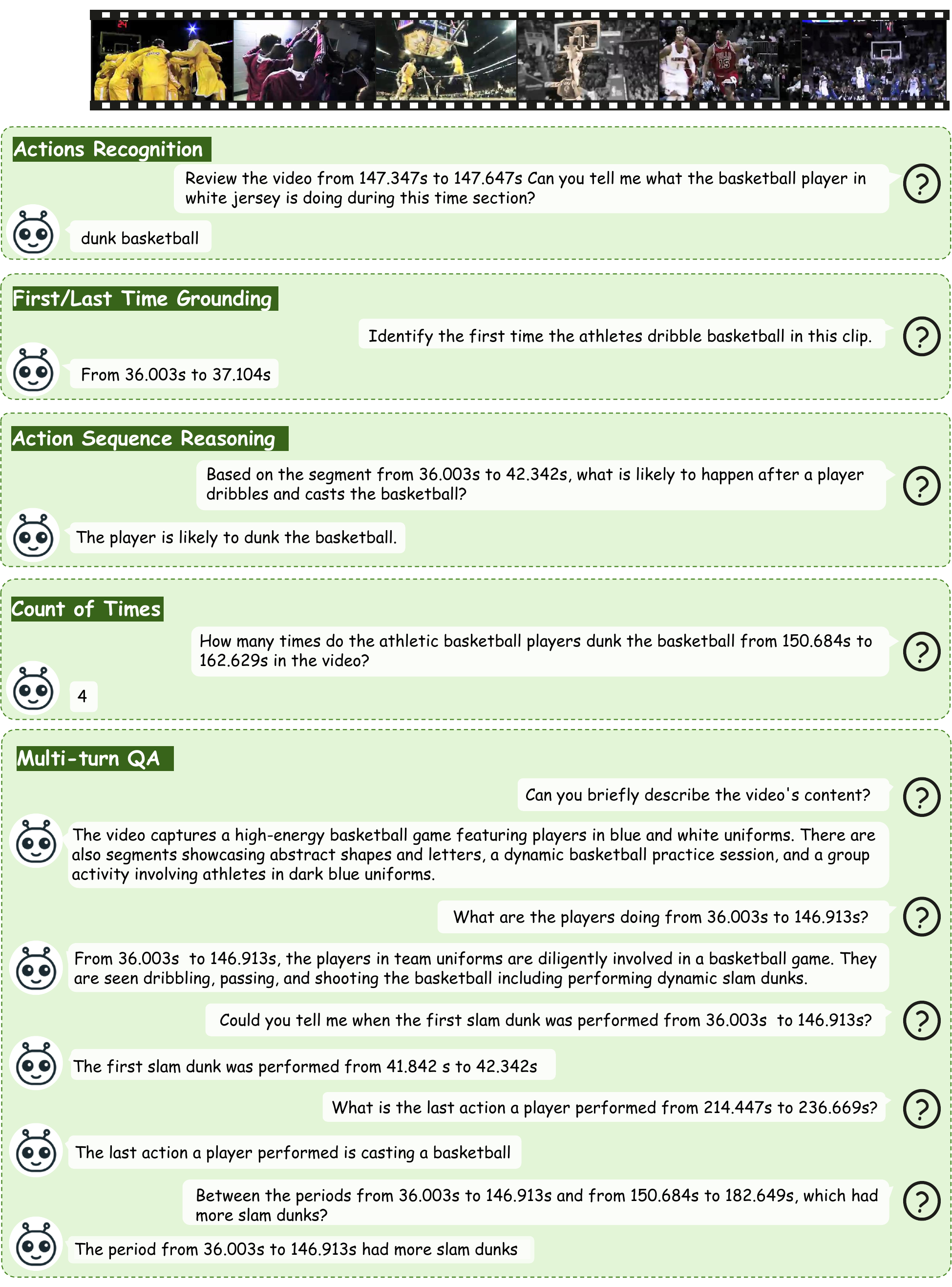}
    \caption{Example of different tasks in FineAction-CGR}
    \label{ap1}
\end{figure}

\begin{figure}[!htp]
    \centering
    \includegraphics[width=1.0\linewidth]{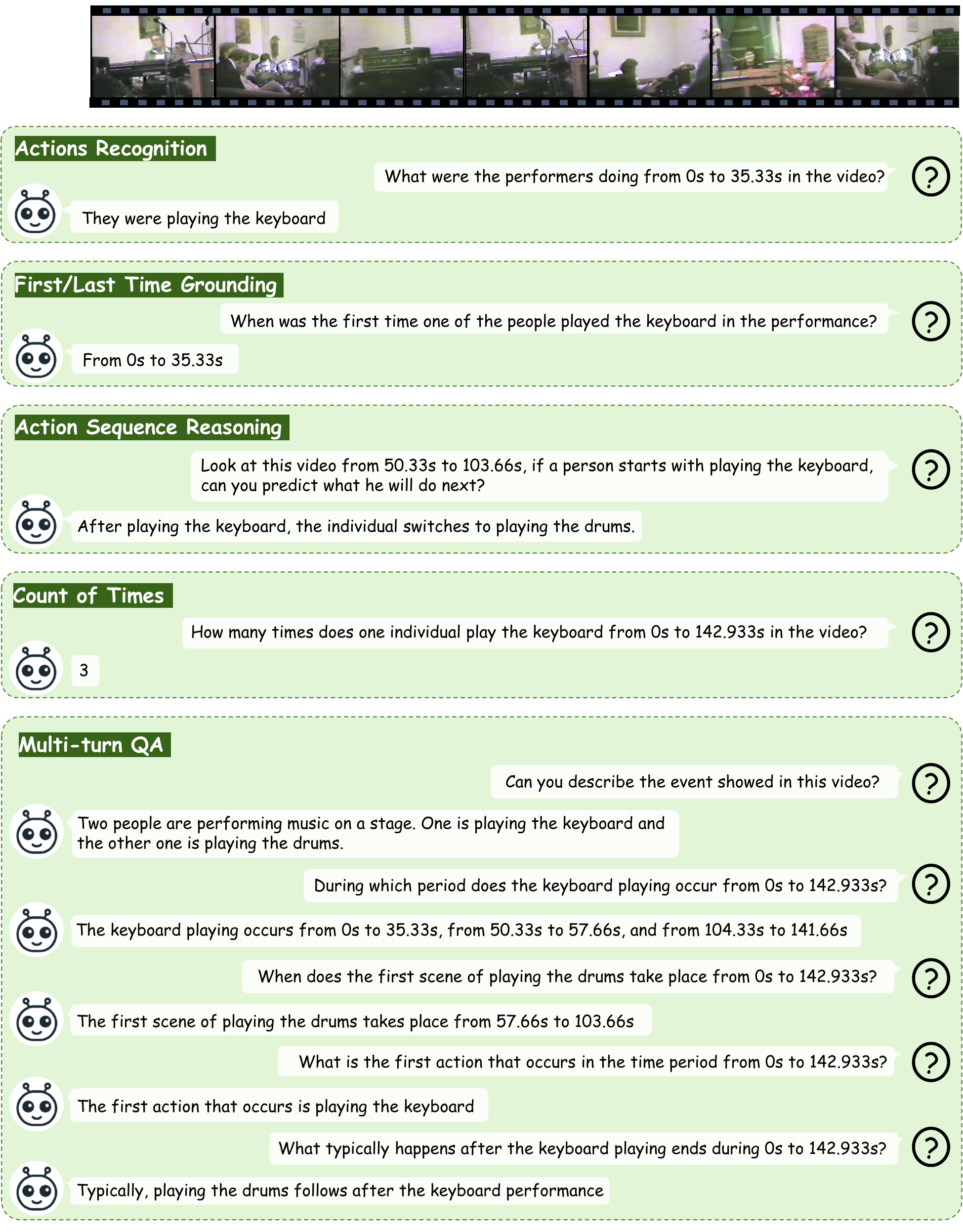}
    \caption{Example of different tasks in FineAction-CGR}
    \label{ap2}
\end{figure}

\textbf{Step III: formation of downstream tasks.} We design various types of instruction-following tasks with different prompts to comprehensively train and evaluate Vid-LLMs. FineAction-CGR encompasses 4 video tasks: (1) captioning, (2) temporal video grounding, (3) temporal video reasoning and (4) multi-turn QA. A few examples from different tasks are shown in Figure~\ref{ap1}-\ref{ap2}.

\textbf{Captioning task} needs model to detects what actions occur in given period.
\begin{itemize}
    \item Action recognition: recognize action in level of clip which may contain a kind of action or several kinds of actions.
\end{itemize}
\textbf{Temporal grounding tasks} aims to predict the boundary in the video given an instruction in which the start and end time of clip are mentioned to attract model's attention to target segment. We design two types of question to achieve fine-grained temporal localization.
\begin{itemize}
    \item First/last time grounding: identify the fist/last time a certain action appears.
\end{itemize}
\textbf{Temporal reasoning tasks} requires model to understand relationship between actions.
\begin{itemize}
    \item Action sequence reasoning: infer possible action before/after an action over a period of time. 
\end{itemize}
\begin{itemize}
    \item Count of times: counts times a certain action appears in a clip.
\end{itemize}
\textbf{Multi-turn QA task} contains multiple rounds of dialogues involving the above three tasks. The problems are progressive from the entire video to the sub-segment.

\begin{figure}[!htp]
    \begin{minipage}{0.5\linewidth}
        \centering
        \includegraphics[height=4cm,width=4cm]{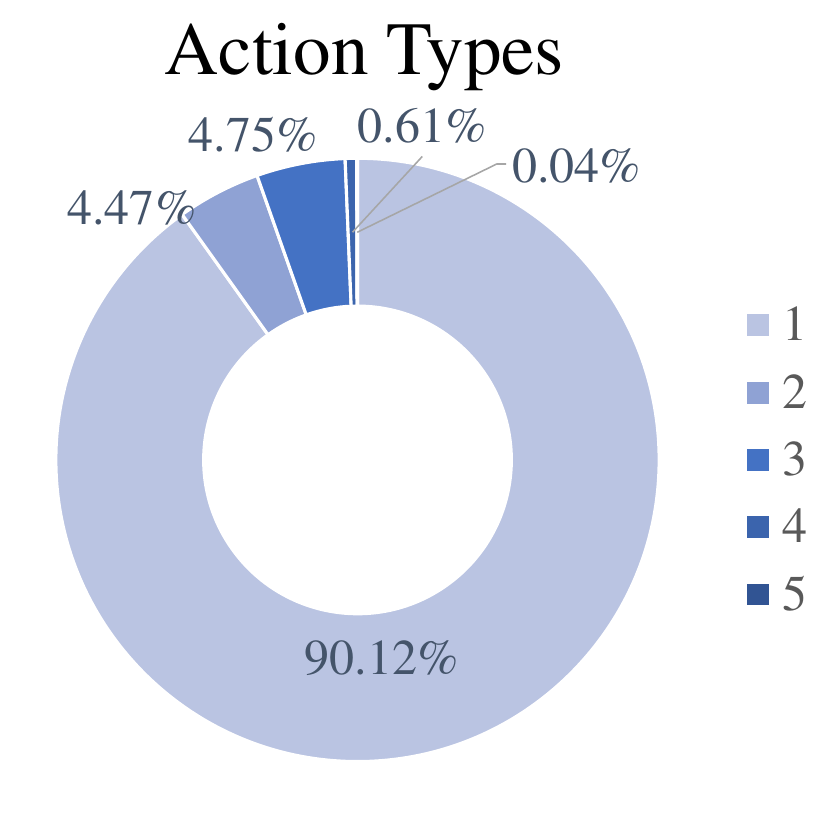}
        \label{fig:action}
    \end{minipage}%
    \begin{minipage}{0.5\linewidth}
        \centering
        \includegraphics[height = 4cm,width=4.3cm]{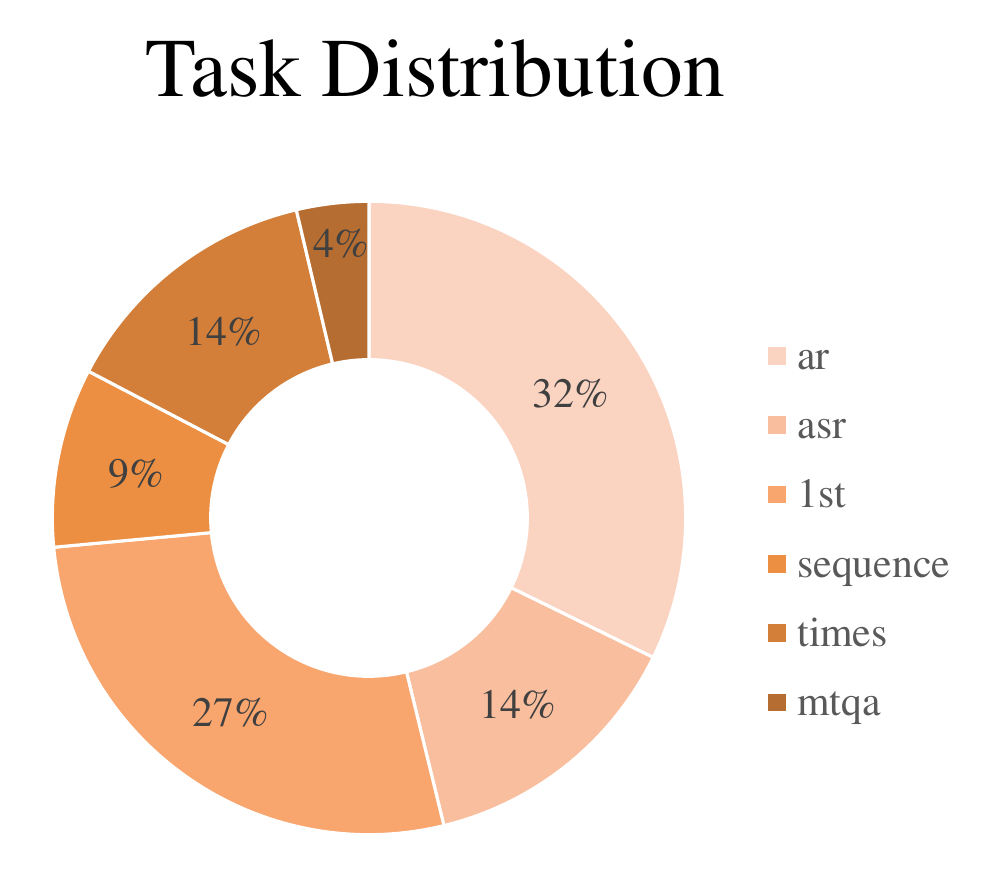}
        \label{fig:task}
    \end{minipage}
    \caption{Video Statistics in FineAction-CGR. It contains a diverse distribution of action types and tasks}
    \label{fig:statistics}
\end{figure}

\begin{figure*}[!htpb]
    \centering
    \includegraphics[width=0.8\linewidth]{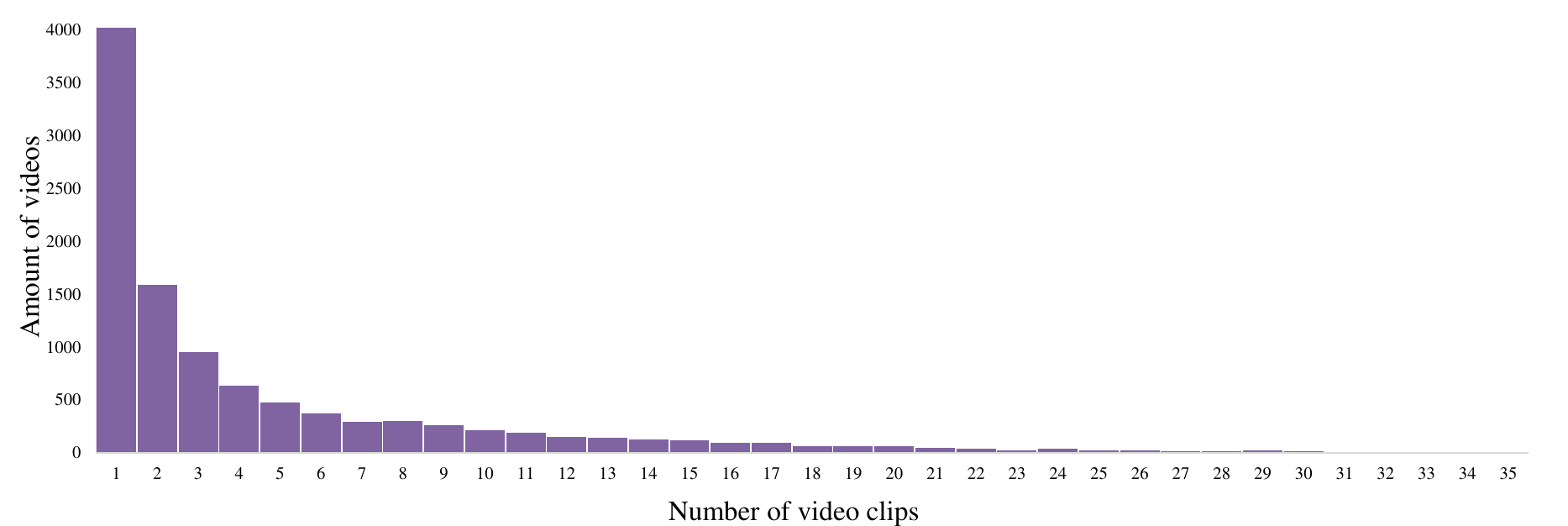}
    \caption{Distribution of clips number in FineAction-CGR}
    \label{fig:clip}
\end{figure*}


\section{Data statistics}\label{app:c}
After filtering, we obtained 11,188 videos with different number of action types. The distribution is illustrated in Figure~\ref{fig:statistics} left. After splitting, videos are segmented into clips. The amount of videos with different clip number is presented in Figure~\ref{fig:clip}, exhibiting a long-tailed distribution. The largest clip number is more than 68. The horizontal coordinate in the figure is only displayed to 35. After generation step, we obtain 131,984 pairs of QA. The distribution in different tasks are illustrated in Figure~\ref{fig:statistics} right. Explanation of the legend 1) 'ar': Captioning task where the model needs to recognize a single action in a clip, 2) 'asr': Captioning task where the model needs to recognize multiple actions in a clip, 3) '1st': Temporal grounding task where the model needs to localize the first/last occurrence of a given action in a clip, 4) 'sequence': Temporal reasoning task where the model needs to identify what action happens before or after a given action in a clip, 5) 'times': Temporal reasoning task where the model needs to count the occurrences of a given action in a clip, 6) 'mtqa': Multi-turn QA task where the model engages in a multi-turn question and answer dialogue about the entire video.

\section{Prompt}\label{app:d}
\begin{figure}[!htp]
\centering
\includegraphics[width=1.0\linewidth]{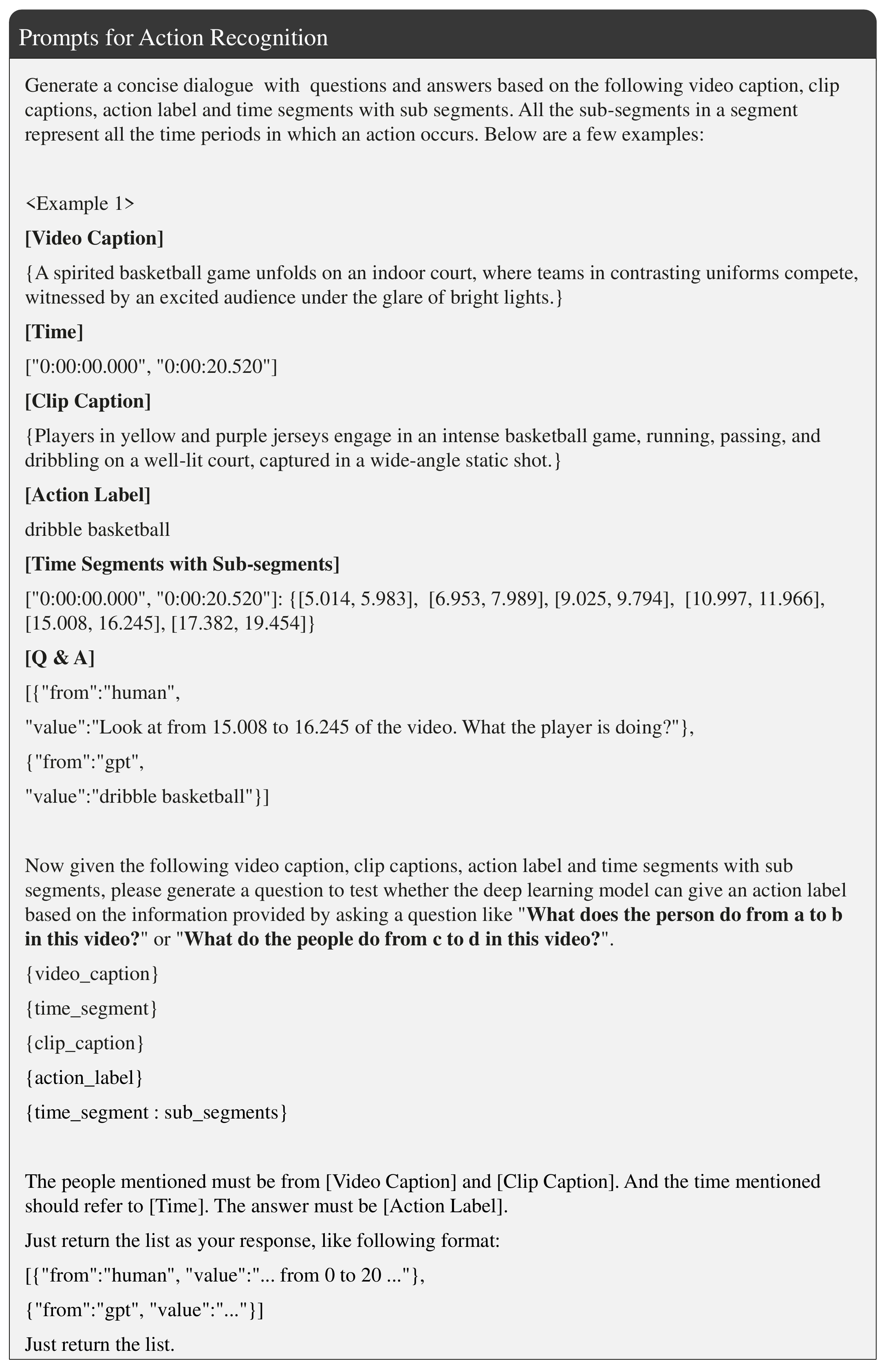}
\caption{Prompt for Action Recognition}
\label{ar}
\end{figure}

\begin{figure}[!htp]
\centering
\includegraphics[width=1.0\linewidth]{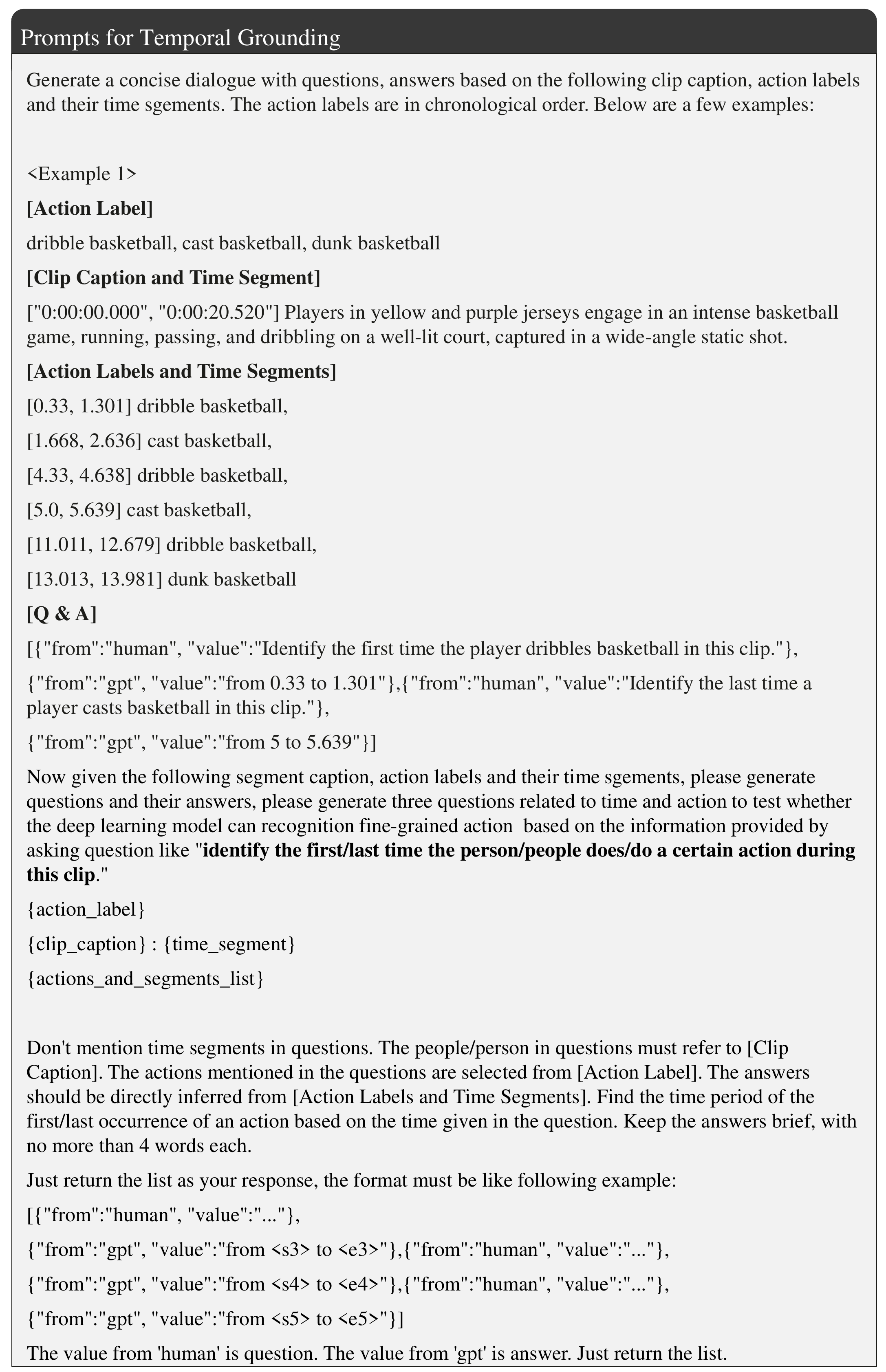}
\caption{Prompt for Temporal Grounding}
\label{tg}
\end{figure}

\begin{figure}[!htp]
\centering
\includegraphics[width=1.0\linewidth]{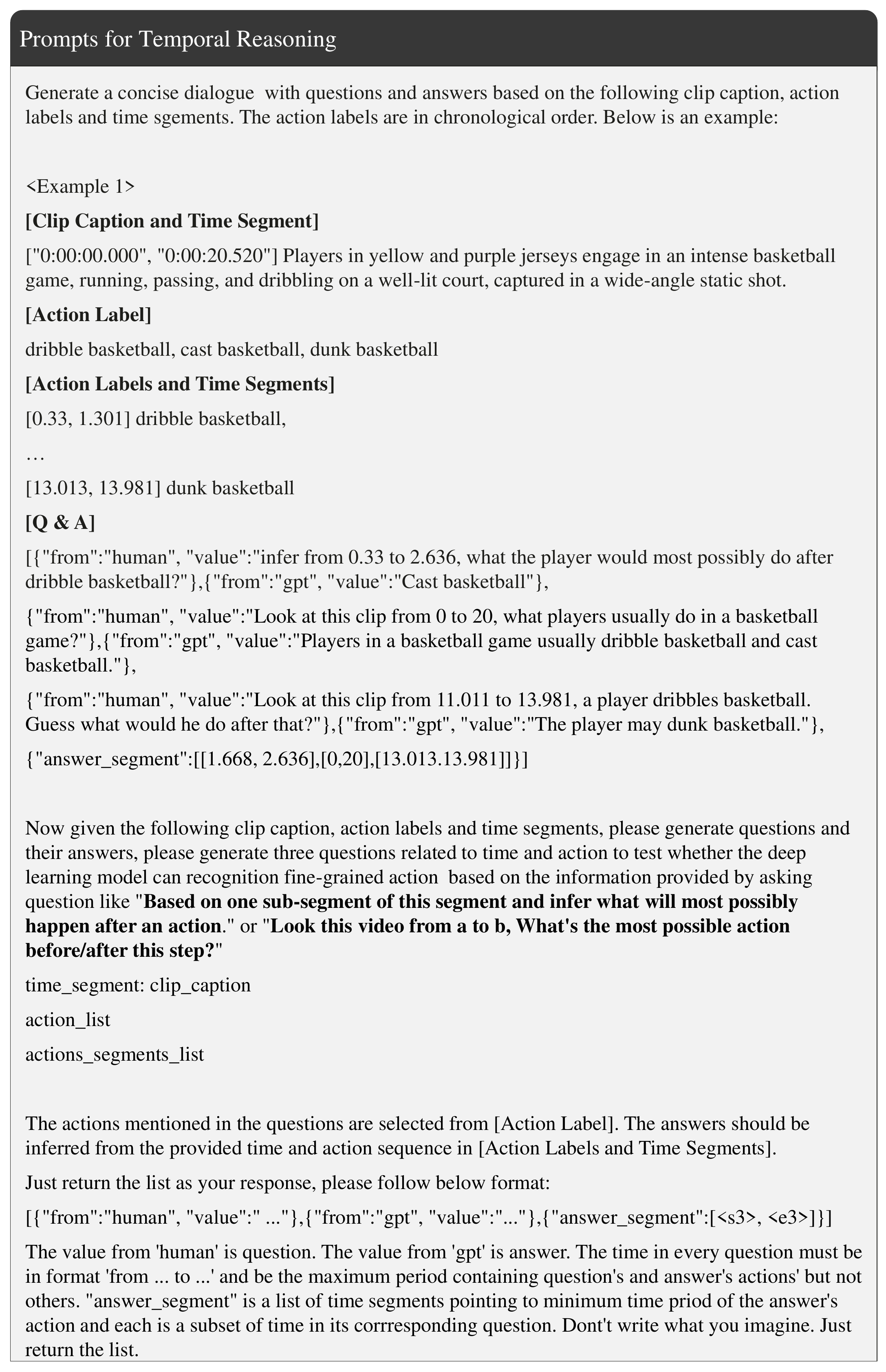}
\caption{Prompt for Temporal Reasoning}
\label{rsn}
\end{figure}

\begin{figure}[!htp]
\centering
\includegraphics[width=1.0\linewidth]{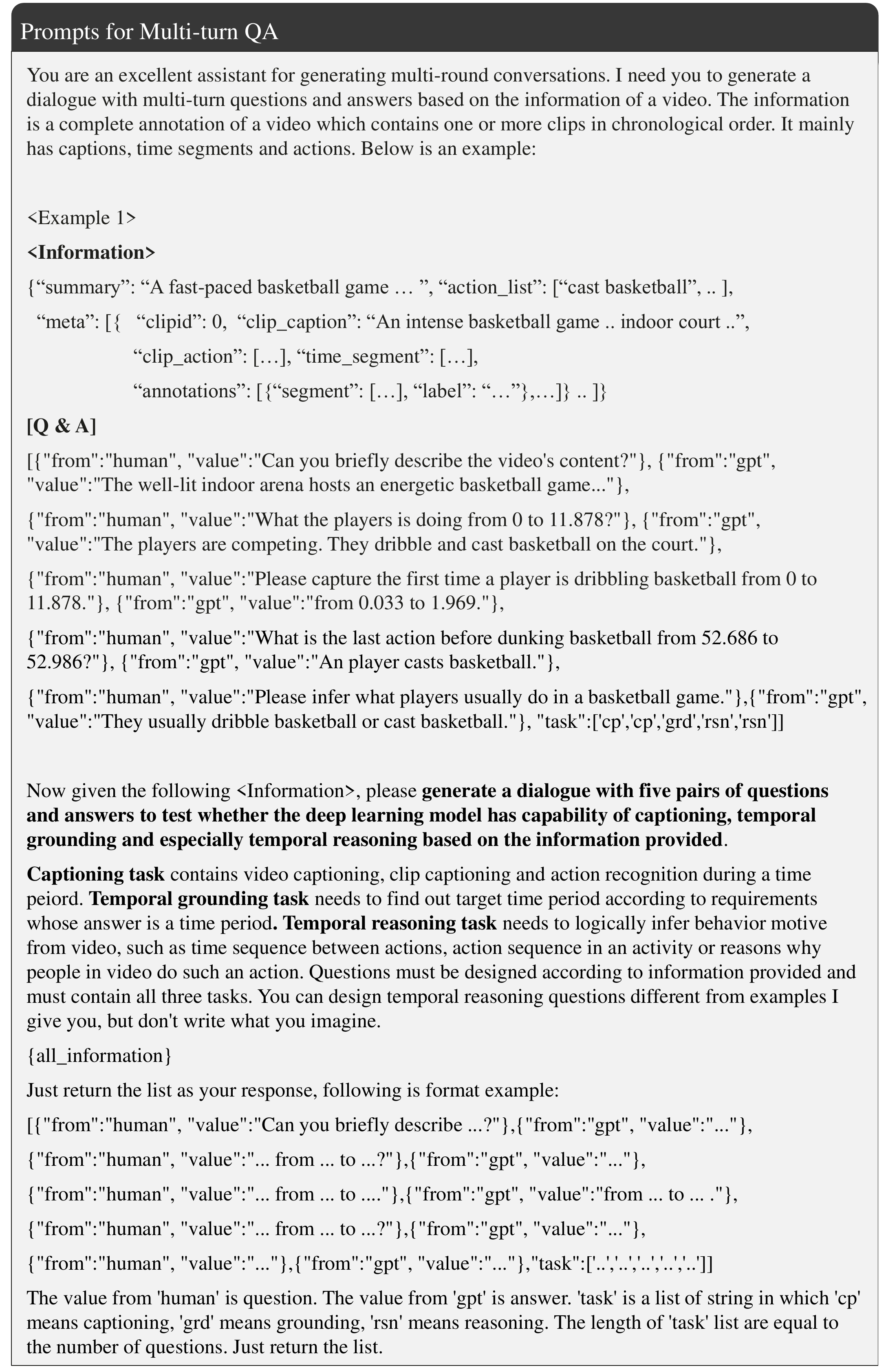}
\caption{Prompt for Multi-turn QA}
\label{qa}
\end{figure}

Prompts used for generation of different kinds of instruction data are shown in Figure \ref{ar}-\ref{qa}. Due to page length constraints, we have omitted some in-context examples in certain tasks.

\section{Broader impacts}
\label{appendix:impacts}
The proposed SlowFocus has the potential to greatly enhance the capability of Video LLMs for video content analysis. Therefore it could be beneficial for applications such as video analytics, surveillance and automated content moderation. It could also be used in educational settings to analyze educational videos to provide a better understanding of complex contents in videos for educational purposes. 

As for the potential negative impacts, there is a risk that the technology could be misused to analyze private videos and spreading disinformation.

\end{document}